\documentclass[10pt,twocolumn,letterpaper]{article}

\usepackage[pagenumbers]{cvpr} % To force page numbers, e.g. for an arXiv version

\usepackage{array}
\usepackage{makecell}
\usepackage{multirow}
\usepackage[table]{xcolor}
\usepackage{colortbl}
\usepackage{graphicx}
\usepackage{pifont}
\usepackage{comment}

\usepackage{fixmetodonotes}

\newif\ifshowedits

\newcommand{\addeditor}[3]{%
  \definecolor{#1color}{rgb}{#3}
  \expandafter\newcommand\csname #1\endcsname[1]{%
  \ifshowedits
    {\color{#1color} ##1}%
  \else
    {##1}%
  \fi
  }%
  \expandafter\newcommand\csname #1rmk\endcsname[1]{%
  \ifshowedits
    {\color{#1color} {\bf [#2: ##1]}}
  \fi
  }%
  \expandafter\newcommand\csname #1rpl\endcsname[2]{%
  \ifshowedits
    {\color{#1color} ##1 \sout{##2}}
  \else
    {##1}
  \fi
  }%
}

\addeditor{alex}{AB}{0.5, 0.5, 0.0}
\addeditor{fatima}{FB}{0, 0.5, 0.5}
\addeditor{raoul}{RC}{.5,.3,.8}

\newcommand{\edit}[1]{{\color{orange}#1}}
\renewcommand{\edit}[1]{#1}

\newcommand{\best}[1]{\textbf{#1}}

\newcommand{\rowOurs}{\rowcolor{red!10}}

\newcommand{\method}{EditSSC}

\showeditsfalse
\newcommand{\todo}[1]{}

\definecolor{cvprblue}{rgb}{0.21,0.49,0.74}
\usepackage[pagebackref,breaklinks,colorlinks,allcolors=cvprblue]{hyperref}

\def\paperID{*****} % *** Enter the Paper ID here
\def\confName{CVPR}
\def\confYear{2026}

\title{\method: Toward Editable Semantic Occupancy Scenes \\ with Unconditional Diffusion Models}

\author{
Fatima Baldé$^{1}$ \quad Raoul de Charette$^{1}$ \quad Alexandre Boulch$^{1,2}$\\
$^{1}$Inria, $^{2}$Valeo.ai\\
}

\begin{document}
\twocolumn[{
\vspace{-2em}
\maketitle
\centering
    \vspace{-1.1em}
    \textbf{\url{https://astra-vision.github.io/EditSSC}}\\[1em]
    \hfill\includegraphics[width=\linewidth]{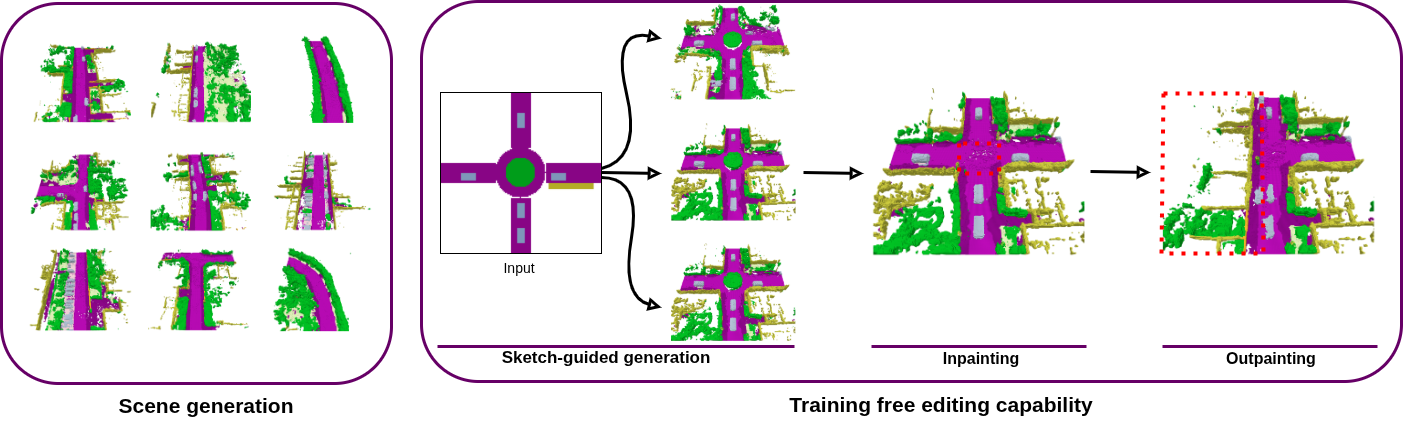}\hfill
    \vspace{-0.9em}
    \captionof{figure}{
    \textbf{\method{} capabilities.} Our scene generation method relies on a latent diffusion model with carefully designed components to enable \textit{training-free} editing capability. While unconditional scene generation (left) can produce multiple diverse samples, all generated scenes faithfully follow the training distribution. \method{} further enables editing via sketch guidance (using a user-provided layout), inpainting, and outpainting. We illustrate these capabilities (right) through a sequential editing process starting from the generation of a rare heavy-traffic roundabout scene, which is then edited by removing the roundabout region and inpainting it (\cf, red highlight). The scene is subsequently outpainted to extend the left portion of the layout. These editing capabilities enable more diverse and controllable scene generation, going beyond what the training data alone provides.  
    }
    \label{fig:teaser}
\vspace{8pt}
}]

\begin{abstract}

3D semantic scene generation is crucial for autonomous driving applications, yet most methods rely on complex 3D-specific architectures such as triplane encoders and adapted diffusion networks, limiting both their simplicity and their editing capabilities. We propose \method{}, an editing-ready method for 3D semantic scene generation using 2D Bird's Eye View (BEV) representations and off-the-shelf latent diffusion network. 
Our approach reshapes 3D semantic occupancy grids into multi-channel BEV images and leverages the quantized autoencoder and UNet from Stable Diffusion with minimal modifications. We perform diffusion on the latents after quantization, which enables training-free editing capabilities. By exploiting class-to-code correspondences in the codebook, our method supports sketch-guided generation, inpainting, and outpainting without any retraining. On SemanticKITTI, \method{} outperforms existing 3D-specific baselines on unconditional generation, demonstrating that well-established 2D architectures can be effectively repurposed for 3D scene generation and editing.

\end{abstract}    
\section{Introduction}
\label{sec:intro}

3D scene generation aims to create geometrically and semantically consistent scenes, with applications in data enrichment and on-the-fly simulation for gaming or navigation. With conditioning and editability, the generated distribution can be steered toward desired targets beyond the training data.

Recently, latent diffusion models  have become the dominant framework for high-quality generation, especially in image synthesis~\cite{rombach2022high, esser2024scaling}, and have naturally been extended to 3D  across modalities such as point clouds~\cite{vahdat2022lion}, meshes~\cite{shue20233d}, and voxel grids~\cite{semcity, wu2024blockfusion}. In the context of scene generation, this entails producing multiple objects and surfaces under realistic spatial constraints. While indoor settings involve restricted layouts with complex object arrangements~\cite{tang2024diffuscene}, outdoor generation focuses on large-scale scenes with objects placed on a ground plane. In this paper, we focus on the latter.

Such large-scale generation typically relies on 3D-specific architectures that explicitly model geometric structure, achieving strong results at the cost of increased complexity and design effort. For example, in SemCity~\cite{semcity} and BlockFusion~\cite{wu2024blockfusion}, scenes are encoded as latent triplanes requiring diffusion networks adapted to this representation and resulting in a complex plane-sharing UNet architecture.

In this paper, we study semantic scene occupancy generation \edit{relying on a carefully designed diffusion-based architecture which unlocks novel scene editing capabilities}.
\edit{Such a property is particularly useful for domains where data diversity is limited, such as autonomous driving, as it allows generating} out-of-training-distribution samples, for instance for training more robust models.

In \cref{sec:pilot_study}, we conduct a pilot study to determine the desirable features of an editing-ready pipeline. In line with recent findings advocating for structured latent space with diffusion models~\cite{lee2025latent}, our observations suggest that the organization of the latent space is more crucial for better diffusion than reconstruction performance alone, and that BEV representations are better suited for editing tasks.

\edit{Building on} these observations, we propose a simple and efficient, editing-ready method, coined \textbf{\method{}}. It uses a 2D representation, allowing us to leverage diffusion pipelines developed for image processing. We first adapt with minimal changes the existing quantized autoencoders for images to ingest voxel grids\edit{, applying diffusion to the quantized latents,} and train a lightweight version of the Stable Diffusion UNet for generation in the latent space. 
While this choice yields comparable results on SemanticKITTI for unconditional scene generation, it enables exploiting an overlooked property of vector quantization for 3D generation. Specifically, the discrete codebook allows us to retrieve class prototypes which can be used at inference to condition the model without any retraining nor test-time adaptation. As a result, as highlighted in~\cref{fig:teaser}, \method{} supports sketch-guided scene generation as well as scene editing via inpainting and outpainting.

\noindent{}To summarize, our contributions are the following:
\begin{itemize}
    \item We repurpose standard 2D diffusion pipelines for 3D semantic occupancy scene generation, showing that well-established 2D architectures can be effectively repurposed for this task;
    \item We show that our simple pipeline  achieves competitive performance on unconditional scene generation;
    \item We demonstrate training-free editing capabilities based on our architecture, including sketch-guided generation, inpainting, and outpainting.
\end{itemize}

\section{Related work}
\label{sec:related}

\noindent\textbf{Diffusion models.}
Diffusion models~\cite{ho2020denoising} aim at fitting a %training 
\edit{distribution through an iterative denoising process}.
Given their tremendous performance, diffusion models have been applied to various fields ranging from image generation~\cite{rombach2022high} and texture synthesis~\cite{lopes2025matswap} to autonomous driving~\cite{liao2025diffusiondrive}.

In image processing in particular, diffusion models are employed in various settings for inpainting or outpainting~\cite{lugmayr2022repaint,saharia2022palette}, as well as for text-to-image conditional generation~\cite{saharia2022photorealistic, rombach2022high}.
Performance for both conditional and unconditional generation has drastically improved via representation alignment  with self-supervised models such as DinoV2~\cite{oquab2023dinov2,yu2024repa}, the enforcement of invariance to data transformation~\cite{kouzelis2025eq} or the introduction of more efficient guidance processes, \eg, classifier-free guidance~\cite{ho2022classifier}.

\noindent\textbf{3D scene generation.}
The success of diffusion models for image generation has naturally led to the exploration of 3D object generation with various data representations such as voxel grids~\cite{li2023diffusion, zhou20213d, maruani2025shapeshifter}, meshes~\cite{liu2023meshdiffusion}, implicit functions~\cite{jun2023shap, shim2023diffusion, shue20233d} or point clouds~\cite{luo2021diffusion, vahdat2022lion}. 
More closely related to our work, scene generation aims at producing a 3D arrangement of multiple shapes. 
Owing to its complexity, it is only recently that diffusion models have been employed for such a task. 
As for objects, various scene representations have been explored using diffusion models for indoor~\cite{meng2025lt3sd, bahmani2023cc3d, ju2024diffindscene}, outdoor~\cite{kim2023neuralfield,liu2024pyramid, scenescalediff,semcity} or both~\cite{ren2024xcube, wu2024blockfusion}. 
In particular, SSD~\cite{scenescalediff} and SemCity~\cite{semcity} tackle 3D outdoor generation with triplane representation. 
At the cost of a 3D-specific design, they propose a %efficient 
\edit{triplane autoencoder and a UNet-based diffusion model adapted to process triplanes}. 
These generative processes have also been used to improve performance on Semantic Scene Completion (SSC), %either 
by directly conditioning the generation on a LiDAR input~\cite{diffssc} or by using diffusion models to refine the output of existing SSC methods~\cite{semcity}. 
Such conditioning is however more complex for methods employing triplane representations.

\edit{We discuss more closely the choice of representation and general design choices for the diffusion models.}

\section{Pilot study}
\label{sec:pilot_study}

From~\cref{sec:related} it follows that recent generative SSC methods typically use two-stage diffusion models with scenes encoded as triplane~\cite{semcity,scenescalediff,diffssc,xi2026flowssc} and diffusion in latent space~\cite{semcity,scenescalediff,xi2026flowssc} learned with a (variational) autoencoder. 
As our goal is to generate and edit 3D semantic scenes, it raises two questions, which we investigate below.
\vspace{-1.2em}
\paragraph{What makes a good editing space?}
The ability to control or edit the generated scenes is particularly desirable for applications like autonomous driving where the available scenes come (i) in limited number and (ii) with a limited diversity.
SemanticKITTI~\cite{semantickitti}, the most popular SSC dataset, is no exception since its training set consists of only  10 sequences from the same German neighborhoods.

\edit{Editing generated scenes is not new and was already addressed by SSEditor~\cite{zheng2026sseditor} which relies on categorical masks for scene editing.}
However, their use of triplane masks is not intuitive for editing as the user needs to draw category profiles in all three  directions. Instead, Bird's Eye View (BEV) appears to be a natural choice, since driving scenes are mainly spread along two axes as objects are rarely stacked. This observation suggests that 2D conditioning should be sufficient for good generation. In addition, BEV representations are practically easier and more intuitive to edit than a full 3D scene.

We observe that existing triplane diffusion can be easily adapted to BEV diffusion by simply pooling the feature volume of the encoder along the vertical direction only. We experiment with this using the SemCity~\cite{semcity} architecture. Specifically, we observe that the autoencoder reconstruction performance drops by 4 IoU and 7 mIoU when switching from triplane to BEV representation.
Subsequently, one might be tempted to assess the superiority of triplane representation. Instead, we question the relationship between reconstruction and diffusion performance.

\vspace{-1.2em}
\edit{
\paragraph{Which design for the autoencoder?}
For generative SSC, it is rather common~\cite{xi2026flowssc,scenescalediff} to justify architectural choices based on autoencoder reconstruction performance, as this is significantly easier to obtain than diffusion results.
This follows a somewhat intuitive belief that better reconstruction performance leads to better generative capability.
However, recent work for images~\cite{lee2025latent} 
has shown that the structure of the latent space, in particular its smoothness and regularity, plays a critical role in diffusion quality, sometimes more than reconstruction fidelity.
To investigate whether similar observations hold for 3D scene generation, we train the diffusion stage on the SemCity~\cite{semcity} architecture and report unconditional scene generation performance.

\begin{table}[t]
    \centering
    \resizebox{0.80\linewidth}{!}{%
    \begin{tabular}{lcc|ccc}
         \toprule
         &\multicolumn{2}{c}{Autoencoder} & \multicolumn{3}{c}{Diffusion}\\
         Representation & IoU$\uparrow$ & mIoU$\uparrow$ & FID$\downarrow$ & CKL$\downarrow$ & Prec$\uparrow$ \\
        \midrule
        \midrule
         \multicolumn{3}{l}{\scriptsize{}\textit{SemCity w/ Autoencoder}}\\
         Triplane~\cite{semcity} & 84.84 & 84.65 & 104.1 & \best{0.0936} & 0.0329\\
         BEV & 80.30 & 77.84 & 120.1 & 0.1310 & 0.0453\\
         \midrule
         \multicolumn{3}{l}{\scriptsize{}\textit{SemCity w/ VQ-VAE}}\\
         BEV & 80.10 & 68.35 & \best{97.5} & 0.0968 & \best{0.0249}\\
        \midrule
        \midrule
         \multicolumn{3}{l}{\scriptsize{}\textit{MLP}}\\
         BEV & \best{98.90} & \best{98.50} & 156.9 & 0.0312 & 0.0115\\
         \bottomrule
         \bottomrule
    \end{tabular}%
    }
    \caption{
    \edit{\textbf{Architectures for diffusion}. We report the original SemCity~\cite{semcity} (triplane) along with two BEV variants using the original SemCity autoencoder (top), a VQ-VAE (middle) and a simple MLP (bottom).} Despite achieving near-perfect reconstruction, the MLP yields the worst diffusion performance, while the VQ-VAE achieves the best despite lower reconstruction scores. This confirms that latent space structure matters more than reconstruction fidelity for diffusion quality. \Cf~\cref{sec:exp_metrics} for metrics details.
    }
    \label{tab:ae}
\end{table}

In~\cref{tab:ae} (`SemCity w/ Autoencoder'), the diffusion performance of the above-mentioned variants correlates with the autoencoder performance, \ie, BEV representation performs worse than triplane.
We then replace the original SemCity autoencoder with a vector-quantized VAE (VQ-VAE), using BEV representation.
Results in~\cref{tab:ae} \edit{(`SemCity w/ VQ-VAE`)} reveal a different picture.
While the VQ-VAE reconstruction performance is significantly lower than its AE counterpart (-9.5 mIoU$\uparrow$), the resulting diffusion performance is notably better (-23 FID$\downarrow$).

For further investigation, we additionally train a simple MLP-based autoencoder that flattens each pillar of the voxel grid into a BEV feature vector. The latter achieves near-perfect reconstruction (IoU: 98.9, mIoU: 98.5), yet yields the worst diffusion performance by a large margin (FID: 156.9), as the latent manifold is sparse and irregular.
This confirms that high reconstruction fidelity with an unstructured latent space is detrimental to diffusion~\cite{lee2025latent}.

Altogether, these observations suggest that reconstruction quality alone is not a reliable proxy for generation capability.
Instead, the discrete and compact latent space imposed by vector quantization provides a structured and regular representation, which is more amenable to diffusion modeling.
This is consistent with the findings of~\cite{lee2025latent}, showing that latent smoothness and regularity are key properties for effective latent diffusion.
It follows that the design of the autoencoder should not only target high reconstruction scores, but also consider the structure of the resulting latent space and its amenability to diffusion modeling. }

\section{Method}
\label{sec:method}

\begin{figure*}[t]
    \centering
    \includegraphics[width=1.0\linewidth]{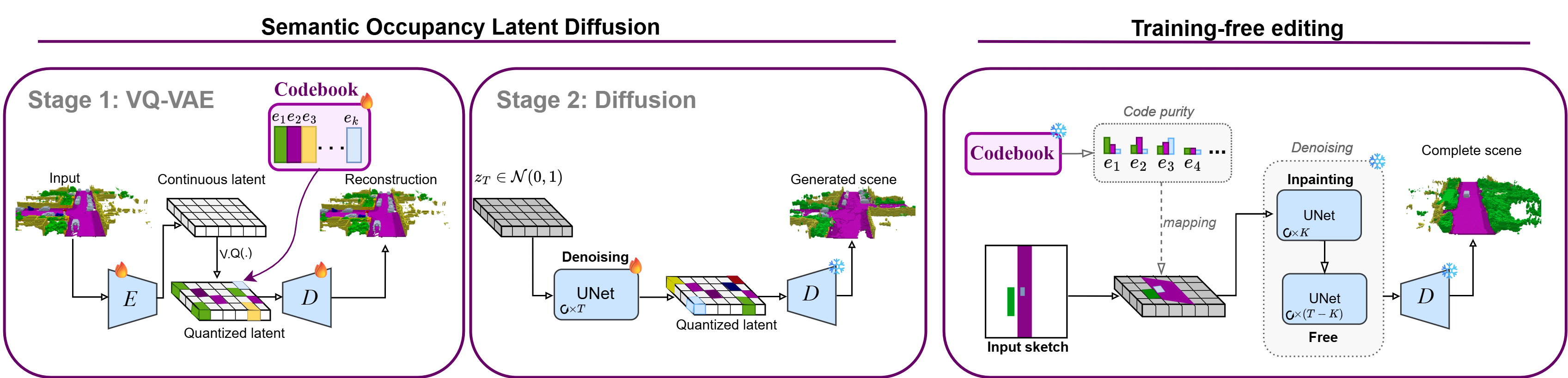}
    \caption{\textbf{Overview of \method{}.} The training consists of two stages. In the first stage (\cref{sec:meth_ae}), a 3D semantic occupancy scene is passed through a VQ-VAE that compresses it into discrete latent codes. In the second stage (\cref{sec:meth_diffusion}), a lightweight U-Net performs diffusion on the quantized latents to generate new scenes. The discrete codebook further enables training-free editing via class-to-code correspondences~(\cref{sec:meth_editing}).}
    \label{fig:pipeline}
\end{figure*}
Our method, coined \method{}, is designed for semantic 3D scene generation with editing capability. As illustrated in~\cref{fig:pipeline}, it relies on a classical two-stage latent diffusion scheme but builds on key observations from~\cref{sec:pilot_study}. First, we use a BEV encoding as it is easier to manipulate. Second, we rely on a vector-quantized autoencoder (VQ-VAE) which has two benefits: (i) it was shown to perform well  with BEV representation (\cf~\cref{tab:ae}), (ii) it has a unique property that enables  training-free in/out-painting. The application of these careful design choices enables editing-ready 3D semantic scene generation.

In detail, as shown in~\cref{fig:pipeline}, in the first stage (\cref{sec:meth_ae}) we train the VQ-VAE from Stable Diffusion~\cite{rombach2022high} to compress 3D semantic occupancy scenes into compact 2D latent representations. The key idea is to reshape the 3D voxel grid into a Bird's Eye View (BEV) image by folding the height dimension into the channel dimension, allowing us to directly leverage a proven image autoencoder without any 3D-specific module. In the second stage (\cref{sec:meth_diffusion}) we train a lightweight version of the Stable Diffusion UNet on the quantized BEV latents to generate new scenes. \edit{Finally, we show in~\cref{sec:meth_editing} that the discrete structure of the VQ-VAE enables training-free editing capabilities.}

\subsection{Autoencoder}
\label{sec:meth_ae}

We consider a 3D semantic occupancy scene represented as a voxel grid of shape $X \times Y \times Z$, where each voxel contains a semantic class label.
To process this 3D input with a 2D image autoencoder, we first map each class label to a learned embedding vector of dimension $D$ using an embedding layer, yielding a tensor of shape $X \times Y \times Z \times D$.
We then reshape this tensor by folding the height and embedding dimensions into the channel axis, obtaining a 2D feature map of shape $X \times Y \times (Z \cdot D)$, which can be interpreted as a multi-channel BEV image encoding the full vertical structure of the scene at each spatial position.
 
This BEV image is passed through the VQ-VAE of Stable Diffusion~\cite{rombach2022high}, by modifying the number of input channels from 3 (RGB) to $Z \cdot D$ and adapting the latent space dimensionality accordingly.
The encoder compresses the BEV image into a spatial grid of discrete latent codes via vector quantization.
At decoding time, the VQ-VAE decoder reconstructs a feature map of shape $X \times Y \times (Z \cdot D)$, which is reshaped back into a 3D volume of shape $X \times Y \times Z \times D$.
A classification head then maps each voxel feature to semantic class logits.
 
Subsequently, the autoencoder jointly trains the embedding layers and classification head in an end-to-end fashion, with the following combined loss:
\begin{equation}
    \mathcal{L}_{\text{VQ-VAE}} = \mathcal{L}_{\text{CE}} + \mathcal{L}_{\text{Lov\'{a}sz}} + \lambda \mathcal{L}_{\text{quant}}
\end{equation}
where $\mathcal{L}_{\text{CE}}$ is the cross-entropy loss providing per-voxel supervision, $\mathcal{L}_{\text{Lov\'{a}sz}}$ is the Lov\'{a}sz-Softmax loss~\cite{berman2018lovasz} which directly optimizes the intersection-over-union (IoU) metric, improving the handling of underrepresented classes, and $\mathcal{L}_{\text{quant}}$ is the VQ-VAE quantization loss~\cite{vqvae} defined as:
\begin{equation}
    \mathcal{L}_{\text{quant}} = \| \operatorname{sg}[\mathbf{z}_e(\mathbf{x})] - \mathbf{e} \|_2^2 + \beta \| \mathbf{z}_e(\mathbf{x}) - \operatorname{sg}[\mathbf{e}] \|_2^2
\end{equation}
where $\mathbf{z}_e(\mathbf{x})$ is the encoder output, $\mathbf{e}$ is the nearest codebook entry, $\operatorname{sg}[\cdot]$ denotes the stop-gradient operator, and $\beta$ controls the commitment weight.
$\lambda$ balances the quantization loss with the reconstruction losses.

\subsection{Diffusion}
\label{sec:meth_diffusion}

Having trained the VQ-VAE, we extract the latent representations of all samples in our training set. We then train a denoising diffusion probabilistic model (DDPM)~\cite{ho2020denoising} directly on the latents obtained after quantization, rather than on the continuous latents before quantization, which we found to yield more stable results (Sec \ref{sec:exp_ablation}).
\vspace{-1.2em}
\paragraph{Forward process.}
Given a clean BEV latent $\mathbf{z}_0$, the forward process gradually adds Gaussian noise over $T = 1000$ steps, producing a sequence of increasingly noisy latents $\mathbf{z}_1, \dots, \mathbf{z}_T$.
Each step follows $q(\mathbf{z}_t | \mathbf{z}_{t-1}) = \mathcal{N}(\sqrt{1 - \beta_t}\, \mathbf{z}_{t-1},\; \beta_t \mathbf{I})$, where $\beta_t$ is a variance schedule.
A useful property of this formulation is that one can directly sample $\mathbf{z}_t$ at any timestep as $q(\mathbf{z}_t | \mathbf{z}_0) = \mathcal{N}(\sqrt{\bar{\alpha}_t}\, \mathbf{z}_0,\; (1 - \bar{\alpha}_t)\, \mathbf{I})$, with $\bar{\alpha}_t = \prod_{i=1}^{t} (1 - \beta_i)$.
At $t = T$, the latent is approximately distributed as pure Gaussian noise.
\vspace{-1.2em}
\paragraph{Reverse process.}
A denoising network $D_\phi$ is trained to reverse this corruption.
Following $x_0$-parameterization~\cite{ho2020denoising}, the network directly predicts the clean sample $\mathbf{z}_0$ from a noisy input $\mathbf{z}_t$ and the timestep $t$, by minimizing:
\begin{equation}
    \mathcal{L}_D = \mathbb{E}_{t \sim \mathcal{U}(1, T)} \| \mathbf{z}_0 - D_\phi(\mathbf{z}_t, t) \|_2^2
\end{equation}
 
For $D_\phi$, we use the UNet architecture from Stable Diffusion~\cite{rombach2022high} retaining attention layers only at the lowest resolution level before the bottleneck. This results in a significantly lighter model, suited to the lower complexity of BEV latent maps compared to natural images.
 
At inference, we start from $\mathbf{z}_T \sim \mathcal{N}(\mathbf{0}, \mathbf{I})$ and iteratively denoise following the DDPM reverse process.
The generated latent $\mathbf{z}_0$ is then decoded through the VQ-VAE decoder and classification head to produce a 3D semantic occupancy scene.

\subsection{Training-free editing}
\label{sec:meth_editing}

An interesting property of our quantized latent space is that the codebook entries exhibit strong correspondences with semantic classes, as illustrated in~\cref{fig:pipeline} (right).
To quantify this, we use the purity of a codebook entry, defined as the proportion of voxels assigned to that entry that belong to its most frequent class.
Based on the training set, we observe that most codebook entries achieve high purity, \ie, they tend to be mostly associated with a single semantic class.
Subsequently, we build a class-to-code mapping by selecting for each class the codebook entry that is both most frequently used by that class and has high purity.
In practice, the two criteria are highly correlated.
This mapping enables training-free editing capabilities without any retraining nor test-time adaptation.

\paragraph{Sketch guidance.} 
Assuming a user-provided BEV sketch, we convert it into latent codes by leveraging our class-to-code mapping and then perform $T$ denoising steps with layout guidance to complete unknown areas. Inspired by RePaint~\cite{lugmayr2022repaint}, we replace the known regions with the layout codes at each denoising step, but only for the first $K$ steps. For the last $T-K$
 steps, we release the constraint and let the model denoise the full latent freely, allowing it to refine the shape and boundaries of the user-specified objects for better coherence with the surrounding scene.

\paragraph{Inpainting/Outpainting.} 
 The above sketch guidance principles extend to inpainting and outpainting, where the known fraction of the scene is preserved at every denoising step, and the model generates coherent content in the missing (or outer) regions throughout the full $T$ denoising steps.

\section{Experiments}
\label{sec:experiments}

In the following, we evaluate \method{} performance along several axes. We first evaluate the quantitative and qualitative performance of our autoencoder in~\cref{subsec:ae_eval} and unconditional scene generation in~\cref{subsec:Unconditional_gen}, while comparing against variants of SemCity~\cite{semcity}. We then evaluate downstream tasks such as LiDAR-conditioned generation in~\cref{sec:exp_condgen}, comparing to various SSC baselines. \edit{We also present qualitative results of the training-free editing capabilities, including sketch-guided generation, inpainting, and outpainting in~\cref{sec:exp_edition}.} Last, we ablate our design and architecture choices in~\cref{sec:exp_ablation}.

\vspace{-1.2em}
\paragraph{Dataset.}
We conduct all our experiments on SemanticKITTI~\cite{semantickitti}, the reference for SSC.
The dataset includes a collection of 11 large-scale outdoor sequences of LiDAR scans with dense semantic occupancy annotations.
\edit{Scenes are encoded as voxel grids} of size $256 \times 256 \times 32$ with a resolution of $0.2$\,m per voxel, covering a spatial extent of $51.2 \times 51.2 \times 6.4$ meters, with 20 semantic classes.
We train our models on the training split and evaluate the VQ-VAE reconstruction on the validation split.

\paragraph{Metrics.}
\label{sec:exp_metrics}
\vspace{-1.2em}
We evaluate the quality of generated scenes using five metrics.
We compute the Fr\'{e}chet Inception Distance (FID)~\cite{heusel2017gans} and Kernel Inception Distance (KID)~\cite{binkowski2018demystifying} on BEV images obtained by projecting the top semantic class of each scene onto the ground plane, where each class is mapped to its corresponding color.
We also compute Precision and Recall to evaluate the fidelity and diversity of the generated scenes, respectively.
Additionally, we report the Categorical KL divergence (CKL), which measures the KL divergence between the per-class frequency distributions of the training set and the generated set, computed on the full 3D voxel grids.
This metric captures whether the generated scenes preserve the overall class distribution of the training data.
All metrics are computed over 5000 generated samples, ensuring reliable FID estimation.
 
Unlike SemCity~\cite{semcity}, which computes these metrics on frontal view renderings, we evaluate on BEV images, which better capture the spatial layout and overall structure of outdoor scenes. For conditional generation tasks, where ground truth is available, we additionally report the mean Intersection-over-Union (mIoU) and per-class Intersection-over-Union (IoU) \edit{to evaluate the semantic accuracy of the generated scenes.}

\subsection{Autoencoder evaluation.}
\label{subsec:ae_eval}

We first evaluate the reconstruction quality of our VQ-VAE on the SemanticKITTI validation set, and compare it against the SemCity autoencoder in its triplane and BEV variants, as well as the SemCity BEV autoencoder augmented with vector quantization (BEV VQ-VAE).
 
\begin{table}[ht]
    \centering
    \resizebox{1.0\linewidth}{!}{%
    \begin{tabular}{lcc|ccc}
         \toprule
         &\multicolumn{2}{c}{Autoencoder} & \multicolumn{3}{c}{Diffusion}\\
         Method & IoU$\uparrow$ & mIoU$\uparrow$ & KID$\downarrow$ & CKL$\downarrow$ & {Prec$\uparrow$} \\
         \midrule
         SemCity (triplane)~\cite{semcity} & {84.84} & {84.65} & 104.1 & {0.0936} & 0.0329\\
         SemCity (BEV)~\cite{semcity} & 80.30 & 77.84 & 120.1 & 0.1310 & 0.0453\\
         SemCity (BEV VQ-VAE) & 80.10 & 68.35 & {97.5} & 0.0968 & \best{0.0249}\\
         \rowOurs{}\method{} (ours) & 81.90 & 72.20 & \best{84.9} & \best{0.0818} & 0.0362\\
         \bottomrule
    \end{tabular}%
    }
    \caption{\textbf{Reconstruction and diffusion performance.} Extending~\cref{tab:ae}, we report the performance of \method{} alongside variants of SemCity. As discussed in~\cref{sec:pilot_study}, we purposely do not highlight best autoencoder performance, as the structure of the latent space plays a more critical role than reconstruction fidelity for generation quality.}
    \label{tab:ae_ours}
\end{table}

Autoencoder metrics are reported in~\cref{tab:ae_ours}, extending the prior results in~\cref{tab:ae}. As observed in~\cref{sec:pilot_study}, the SemCity triplane autoencoder achieves the highest reconstruction scores, benefiting from its \raoul{rich} three-plane representation.
However, as detailed in~\ref{sec:pilot_study}, this does not translate into better generation performance.
Among quantized architectures, \method{}, which employs a VQ-VAE based on Stable Diffusion, outperforms the SemCity BEV VQ-VAE in both IoU and mIoU, confirming the effectiveness of repurposing a standard 2D autoencoder for 3D semantic occupancy.

\begin{figure}
\centering
\setlength{\tabcolsep}{0.5pt}
\resizebox{\linewidth}{!}{
\scriptsize
\begin{tabular}{cccccc}
\rotatebox{90}{Train samples}&\includegraphics[width=0.19\linewidth]{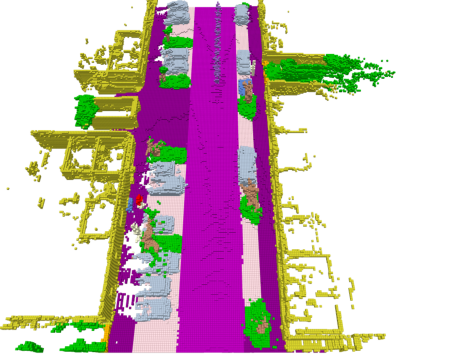} &
\includegraphics[width=0.19\linewidth]{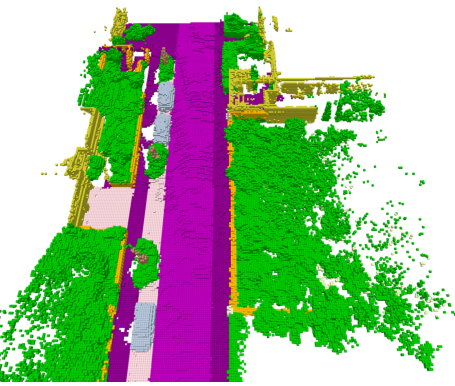} &
\includegraphics[width=0.19\linewidth]{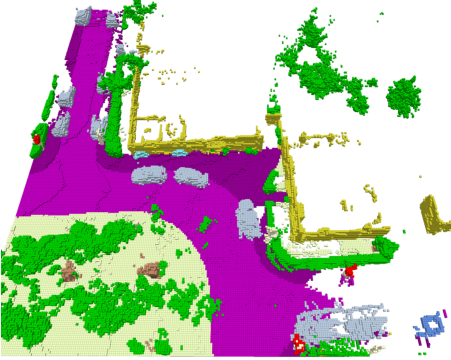} &
\includegraphics[width=0.19\linewidth]{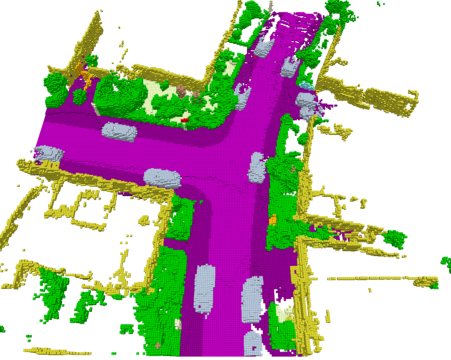} &
\includegraphics[width=0.19\linewidth]{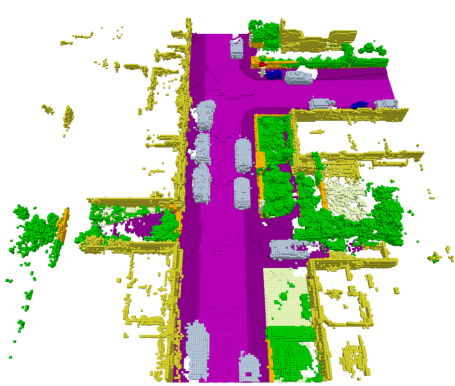} \\
\rotatebox{90}{\hspace{0.0em}\method{} (ours)}&
\includegraphics[width=0.19\linewidth]{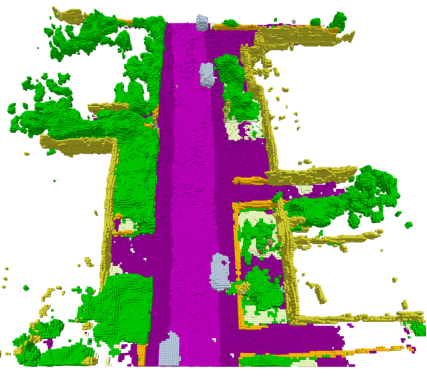} &
\includegraphics[width=0.19\linewidth]{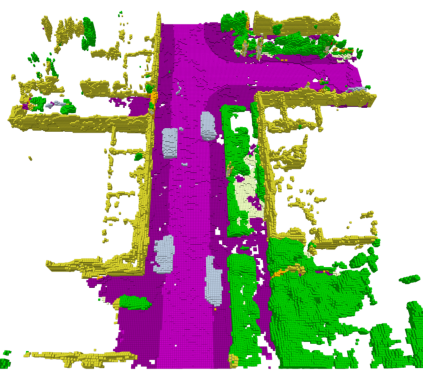} &
\includegraphics[width=0.19\linewidth]{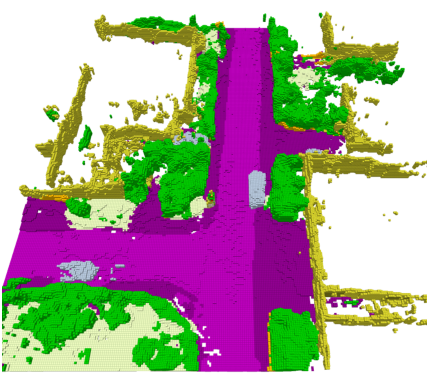} &
\includegraphics[width=0.19\linewidth]{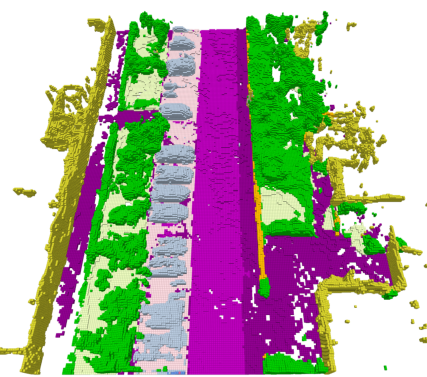} &
\includegraphics[width=0.19\linewidth]{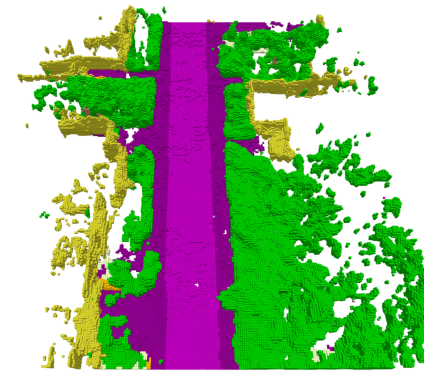} \\
\end{tabular}
}
\caption{\raoul{\textbf{Unconditional generation}. Our method generates plausible scenes (bottom) which follow the class distribution and general structure of the training set (top).}}
\label{fig:qualitative}
\end{figure}

\begin{figure}
\centering
\setlength{\tabcolsep}{1pt}
\begin{tabular}{ccc}
\includegraphics[width=0.3\linewidth]{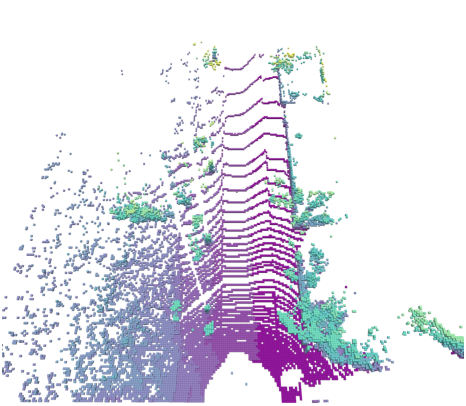} &
\includegraphics[width=0.3\linewidth]{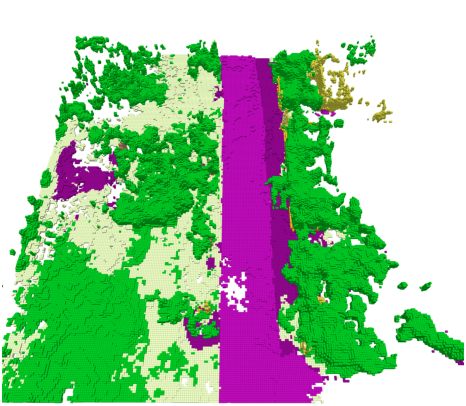} &
\includegraphics[width=0.3\linewidth]{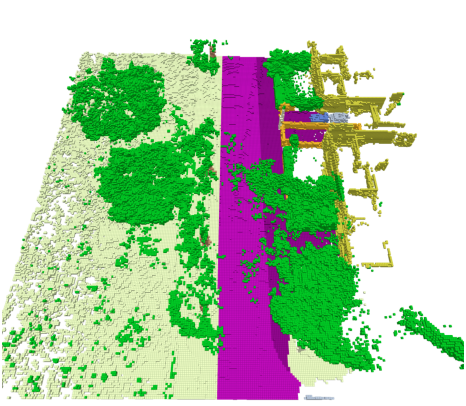} \\[2pt]
\includegraphics[width=0.3\linewidth]
{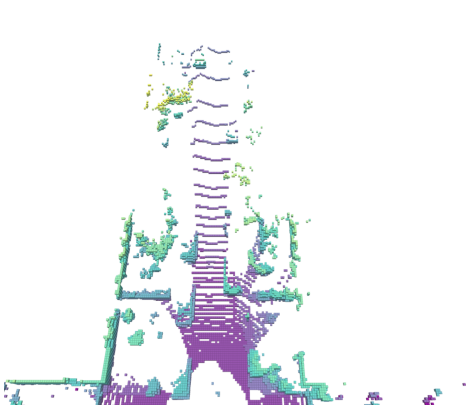} &
\includegraphics[width=0.3\linewidth]{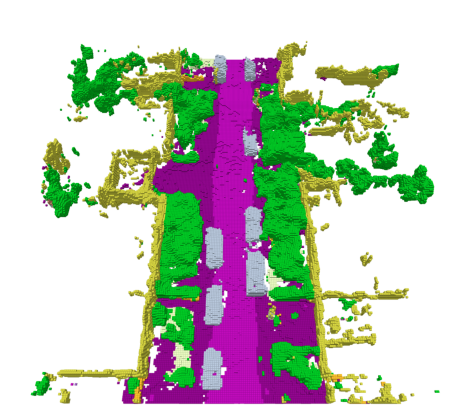} &
\includegraphics[width=0.3\linewidth]{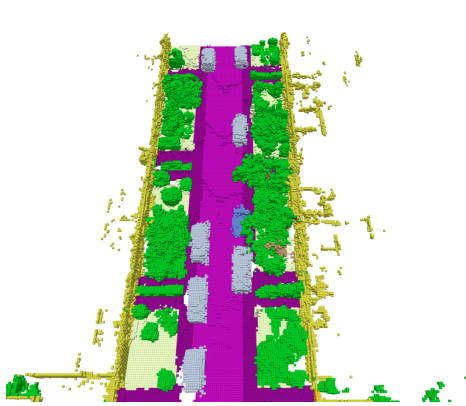} \\[2pt]
\includegraphics[width=0.3\linewidth]
{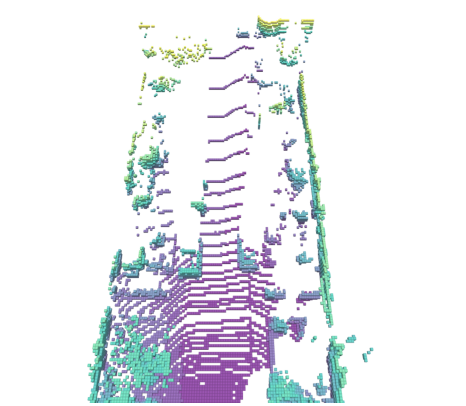} &
\includegraphics[width=0.3\linewidth]{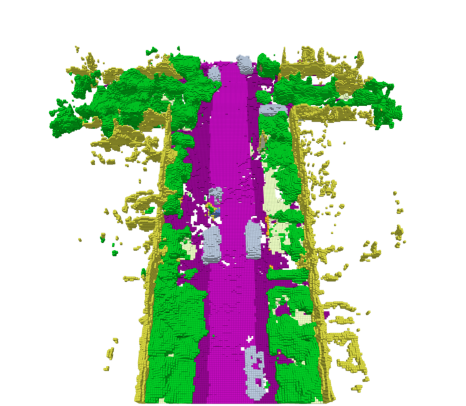} &
\includegraphics[width=0.3\linewidth]{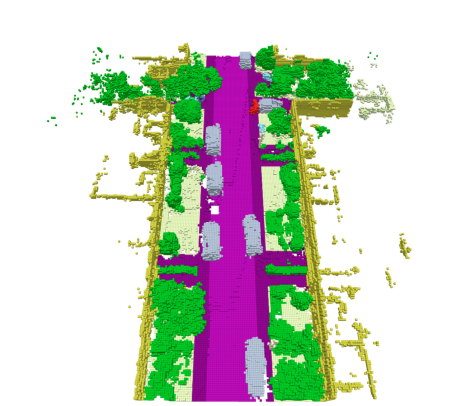} 
\end{tabular}
\caption{\textbf{LiDAR-conditioned generation.} Given a LiDAR scan~(left), our model generates a semantic occupancy scene (middle) \edit{which resembles the ground truth} (right).}
\label{fig:lidar_condition}
\end{figure}

\begin{figure*}[t]
\centering
\setlength{\tabcolsep}{4pt}
\resizebox{0.95\linewidth}{!}{
\begin{tabular}{ccc|ccc}
\textbf{Layout} & \multicolumn{2}{c}{\textbf{Generated Scenes}} & \textbf{Layout} & \multicolumn{2}{c}{\textbf{Generated Scenes}} \\[2pt]
\fbox{\includegraphics[height=3.2cm,width=3.2cm]{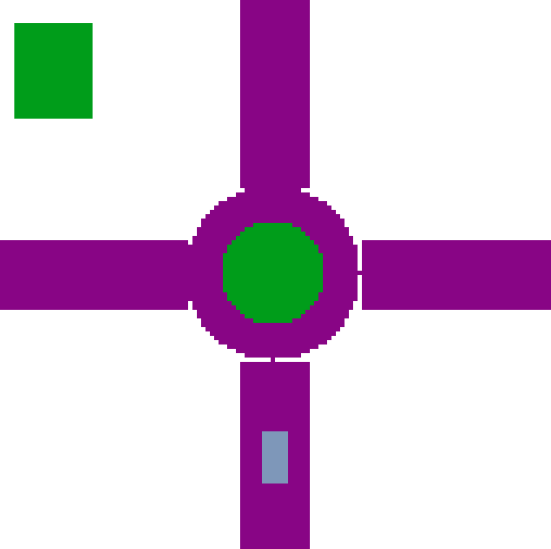}} &
\includegraphics[height=3.2cm]{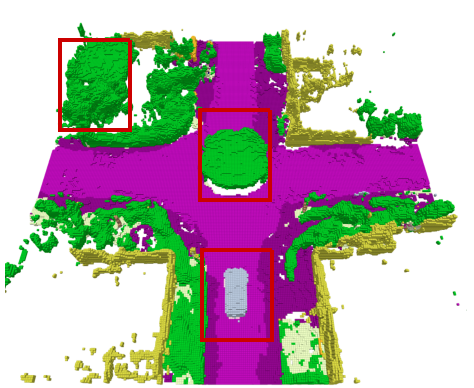} &
\includegraphics[height=3.2cm]{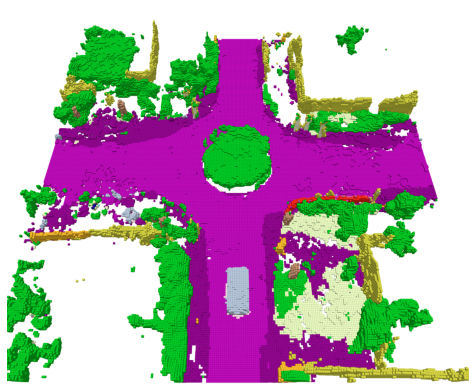} &
\fbox{\includegraphics[height=3.2cm,width=3.2cm]{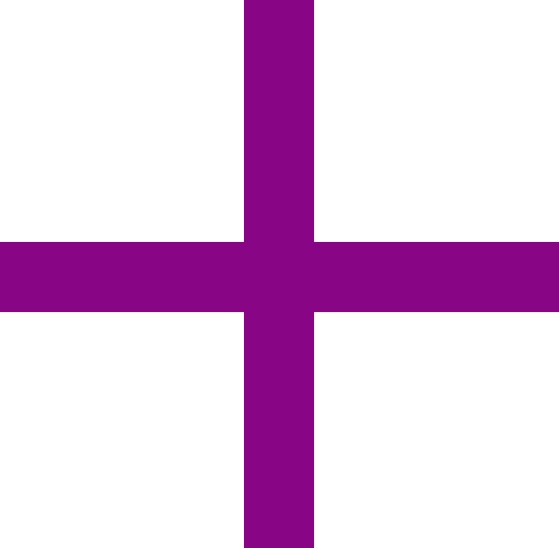}} &
\includegraphics[height=3.2cm]{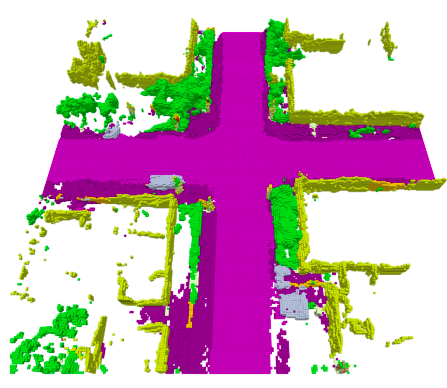} &
\includegraphics[height=3.2cm]{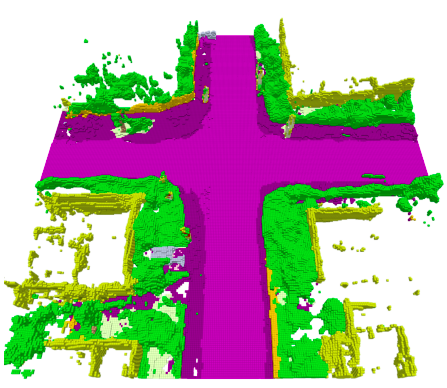} \\[0.5em]
\fbox{\includegraphics[height=3.2cm,width=3.2cm]{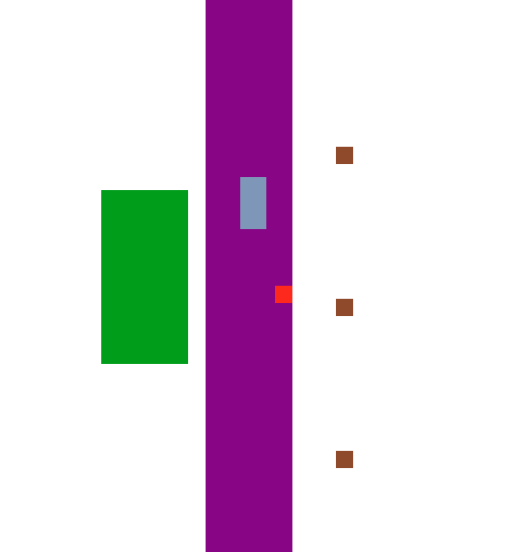}} &
\includegraphics[height=3.2cm]{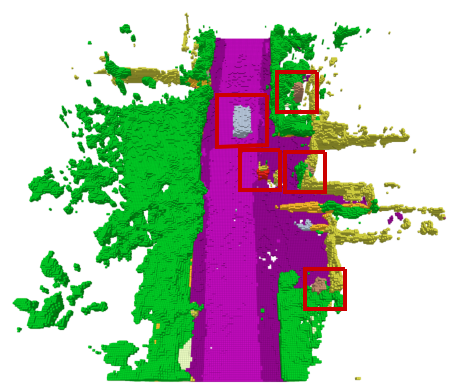} &
\includegraphics[height=3.2cm]{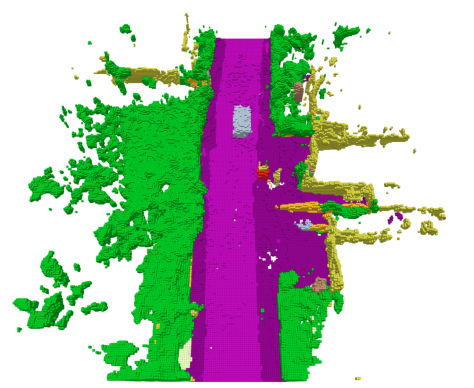} &
\fbox{\includegraphics[height=3.2cm,width=3.2cm]{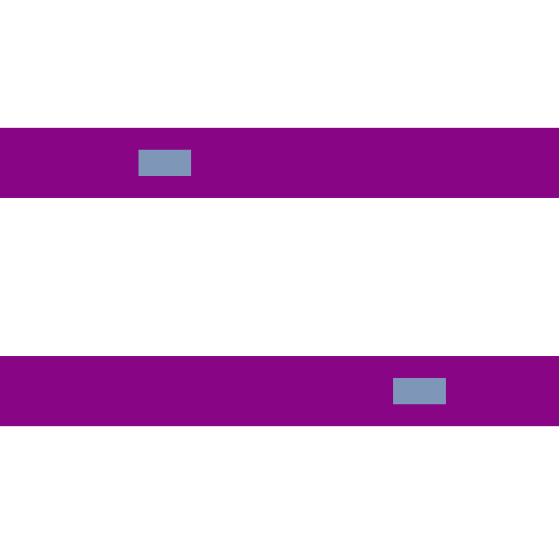}} &
\includegraphics[height=3.2cm]{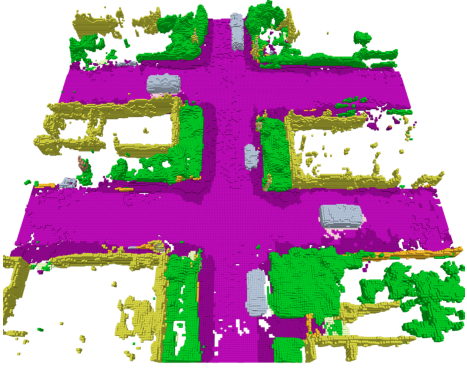} &
\includegraphics[height=3.2cm]{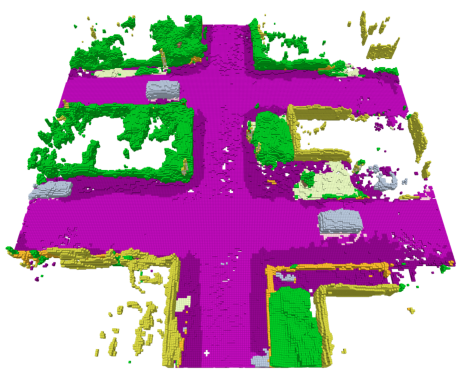} \\[0.5em]
\fbox{\includegraphics[height=3.2cm,width=3.2cm]{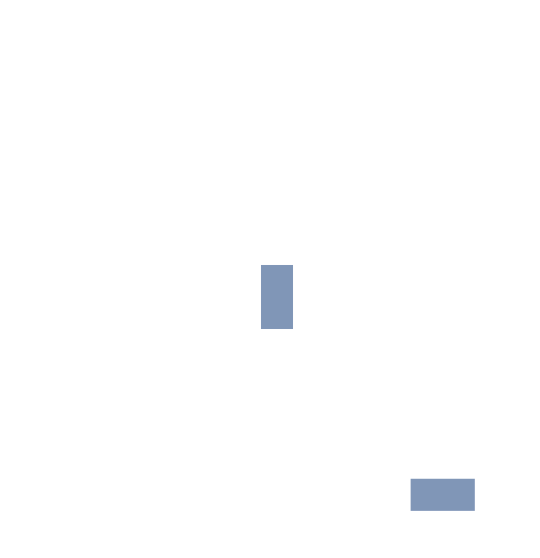}} &
\includegraphics[height=3.2cm]{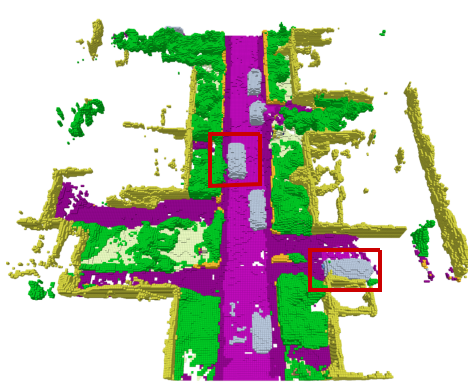} &
\includegraphics[height=3.2cm]{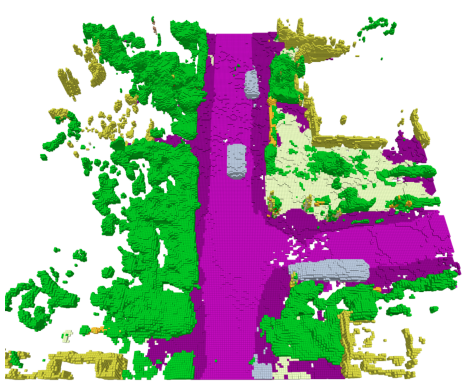} &
\fbox{\includegraphics[height=3.2cm,width=3.2cm]{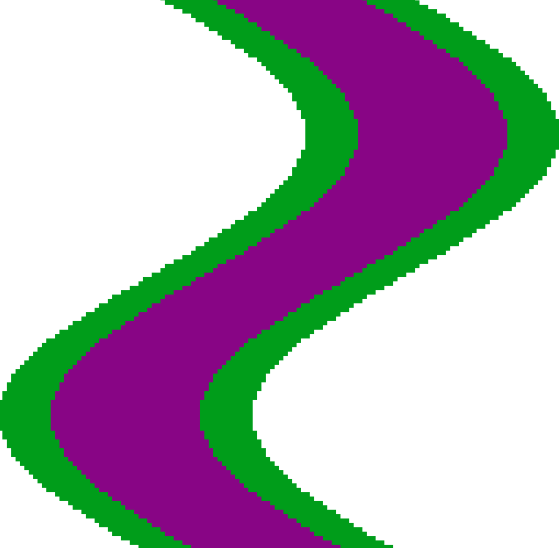}} &
\includegraphics[height=3.2cm]{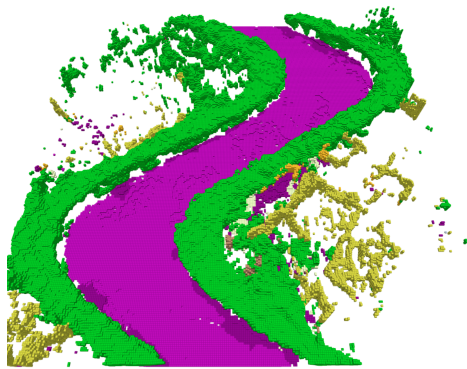} &
\includegraphics[height=3.2cm]{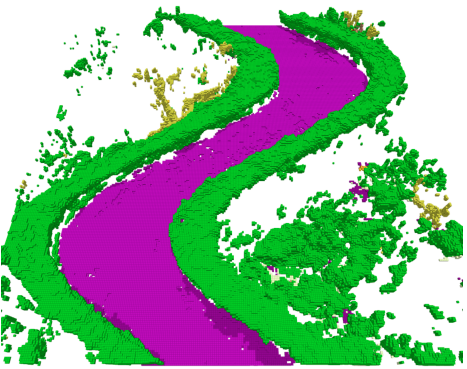} \\[0.5em]
\fbox{\includegraphics[height=3.2cm,width=3.2cm]{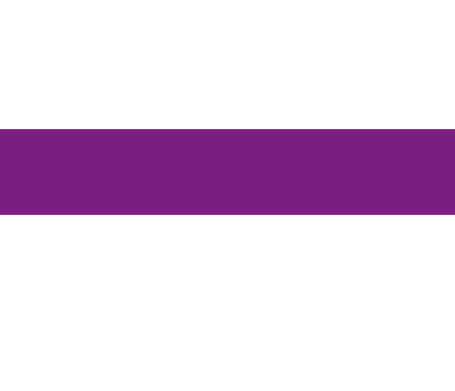}} &
\includegraphics[height=3.2cm]{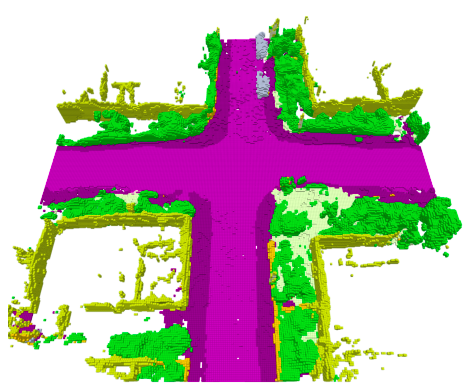} &
\includegraphics[height=3.2cm]{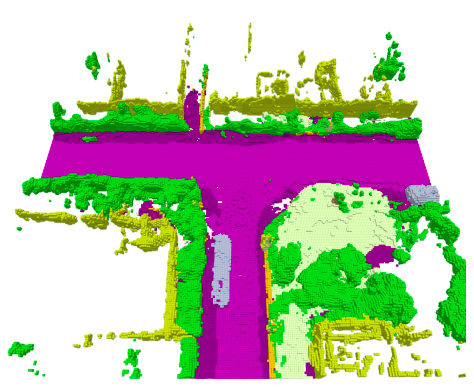} &
\fbox{\includegraphics[height=3.2cm,width=3.2cm]{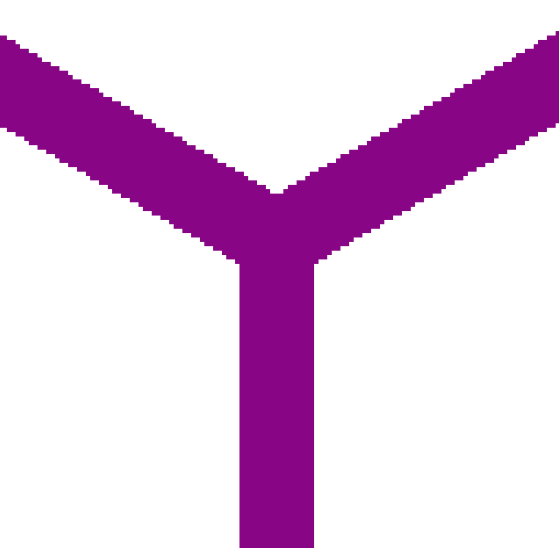}} &
\includegraphics[height=3.2cm]{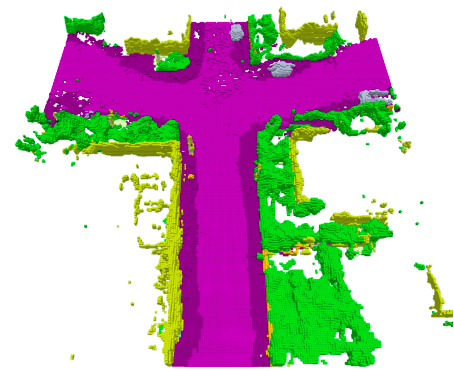} &
\includegraphics[height=3.2cm]{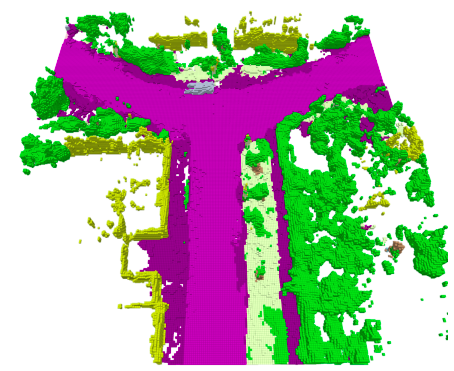} \\
\end{tabular}
}
\caption{\textbf{Training-free sketch-guided generation.} Given a user-drawn BEV layout, our method generates diverse and coherent 3D semantic scenes without any retraining. Each layout is shown with two independently generated scenes, demonstrating both fidelity to the user-specified structure and diversity in the completions.}
\label{fig:training_free}
\end{figure*}

\subsection{Unconditional generation.}
\label{subsec:Unconditional_gen}

For unconditional generation, we sample a random BEV latent $\mathbf{z}_T \sim \mathcal{N}(\mathbf{0}, \mathbf{I})$ and iteratively denoise it using the trained UNet following the DDPM~\cite{ho2020denoising} reverse process.
The resulting latent $\mathbf{z}_0$ is then passed through the VQ-VAE decoder and classification head to reconstruct a full 3D semantic occupancy scene.
 
We report results of both our method and all three variants of SemCity, using 100 denoising steps at inference, with the training set as reference distribution. Diffusion performance in \cref{tab:ae_ours} shows that our fully 2D pipeline outperforms baselines on two out of three metrics, with a large gap in KID (-6.6) and CKL (-0.015) showing that it produces not only more realistic scenes but also more faithfully reproduces the training class distribution.
Notably, even the SemCity BEV VQ-VAE variant, which uses 3D convolutions in its encoder, falls behind our approach based on a standard 2D VQ-VAE.
This further reinforces that 3D-specific modules such as 3D convolutions are not necessary for high-quality outdoor semantic occupancy generation, and that well-established 2D architectures can be effectively repurposed for this task.

In~\cref{fig:qualitative}, we also report examples of scenes generated by our method, alongside reference samples from the training set. Visuals confirm the ability of \method{} to produce realistic scenes, which visually resemble those seen in the training set.

\subsection{Conditional generation.}
\label{sec:exp_condgen}
We also evaluate the ability of \method{} to perform LiDAR-conditioned scene generation, a well-established task.
To do so, we voxelize the input LiDAR scan, which is then passed to an encoder composed of 3D convolutional layers.
The resulting features are then pooled along the height axis to obtain a BEV feature map, which is concatenated with the noisy latent as input to the diffusion model.
The model is trained end-to-end with this conditioning signal.

We report results in~\cref{tab:lidarcond} alongside existing SSC methods\edit{, highlighting} that \method{} is the only truly editable SSC method  \edit{that supports LiDAR-conditioned generation}. \edit{Notably, SSEditor~\cite{zheng2026sseditor} does not support LiDAR-conditioned generation due to its complex architecture}.
Compared to other SSC techniques, we observe a gap in IoU and mIoU, as existing SSC methods largely outperform \method{}, highlighting the need for further research on editable SSC. 
 \edit{Yet, qualitative results in~\cref{fig:lidar_condition} demonstrate that our method faithfully follows the input LiDAR scan geometry~(left) to generate plausible reconstructions~(middle). Besides road layout, note for example that vehicles are accurately positioned in the generations \wrt the input scan.}

\begin{table}
    \centering
    \resizebox{0.55\linewidth}{!}{%
    \begin{tabular}{ccc}
         \toprule
         Method & IoU$\uparrow$ & mIoU$\uparrow$\\
         \midrule
         \multicolumn{3}{l}{\scriptsize{}\textit{SSC}}\\
         TS3D~\cite{ts3d} & 50.6 & 17.7\\
         LMSCNet~\cite{lmscnet} & 55.7 & 17.0\\
         JS3C-Net~\cite{js3cnet} & 57.0 & 24.0\\
         DPS2CNet~\cite{liu5333789dual} & 60.8 & 26.7\\
         DiffSSC~\cite{diffssc} & 60.3  & 26.7\\
         \midrule
         \multicolumn{3}{l}{\scriptsize{}\textit{Editable SSC}}\\
         \rowOurs{}\method{} (ours) & {42.1} & {12.5}\\
         \bottomrule
    \end{tabular}%
    }
    \caption{\raoul{\textbf{LiDAR-conditioned generation on SemanticKITTI.} We report the performance of general SSC methods as well as editable SSC methods on SemanticKITTI (val. set). We intentionally omit SSEditor, which does not report \edit{LiDAR-conditioned performance.}}
    }
    \label{tab:lidarcond}
\end{table}

\subsection{Training-free editing capabilities.}
\label{sec:exp_edition}
\begin{figure}[t]
\centering
\setlength{\tabcolsep}{1pt}
\begin{tabular}{ccc}
\multicolumn{3}{c}{\textbf{Inpainting}} \\[2pt]
\textbf{Masked} & & \textbf{Completed} \\[0pt]
\includegraphics[width=0.3\linewidth]{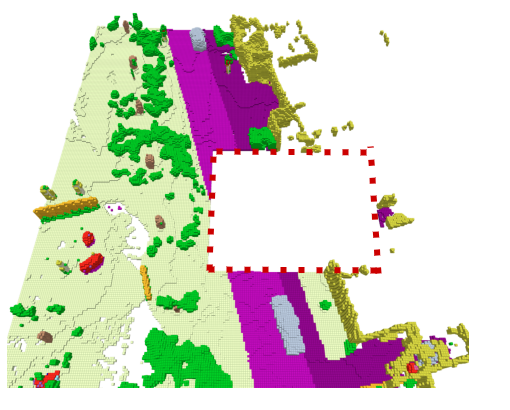} &
\raisebox{6\height}{\large$\boldsymbol{\rightarrow}$} &
\includegraphics[width=0.3\linewidth]{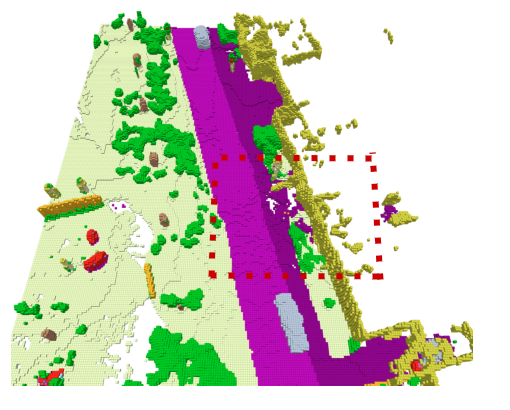} \\[2pt]
\includegraphics[width=0.3\linewidth]{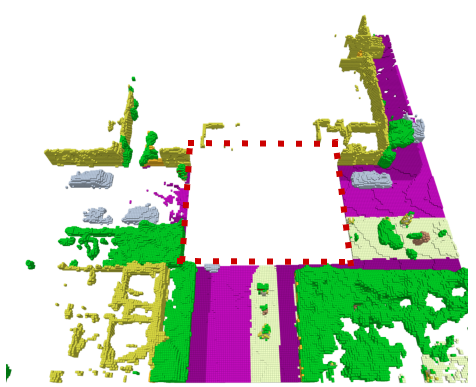} &
\raisebox{6\height}{\large$\boldsymbol{\rightarrow}$} &
\includegraphics[width=0.3\linewidth]{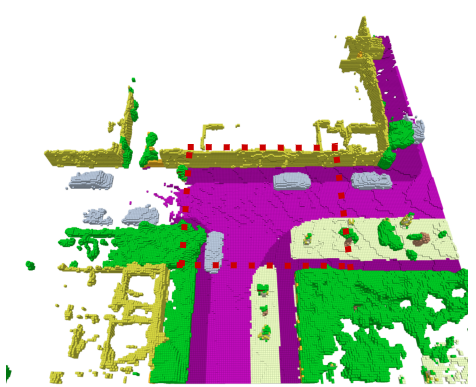} \\[6pt]
\multicolumn{3}{c}{\textbf{Outpainting}} \\[2pt]
\textbf{Input} & & \textbf{Extended} \\[0pt]
\includegraphics[width=0.3\linewidth]{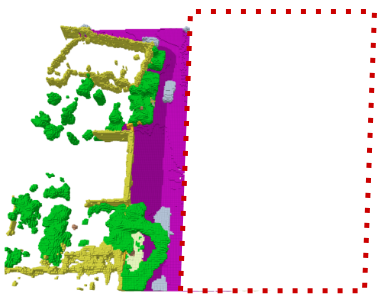} &
\raisebox{6\height}{\large$\boldsymbol{\rightarrow}$} &
\includegraphics[width=0.3\linewidth]{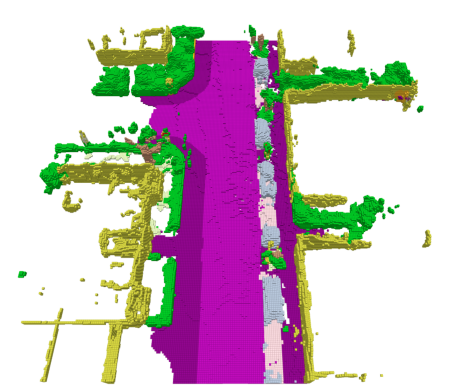} \\[2pt]
\includegraphics[width=0.3\linewidth]{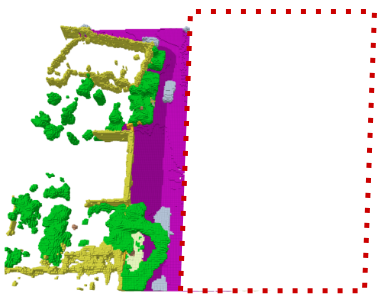} &
\raisebox{6\height}{\large$\boldsymbol{\rightarrow}$} &
\includegraphics[width=0.3\linewidth]{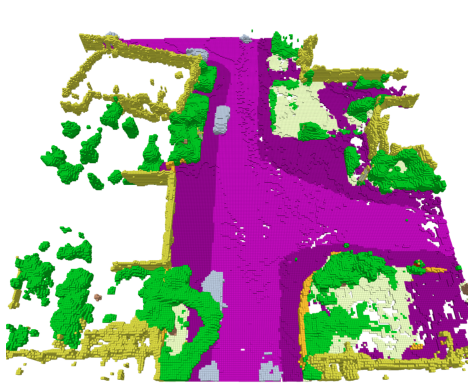} \\
\end{tabular}
\caption{\textbf{Scene inpainting and outpainting.} Inpainting (top): given a scene with a masked region, our model generates coherent content to fill the missing area while preserving the known regions. Outpainting (bottom): given half of an existing scene, our model extends it by generating the missing half, producing a spatially coherent continuation.}
\label{fig:inpainting_outpainting}
\end{figure}

\edit{
We present qualitative results of the training-free editing capabilities introduced in~\cref{sec:meth_editing}.

\paragraph{Sketch-guided generation.}

We illustrate the sketch-guided generation capability in~\cref{fig:training_free}, using simple hand-drawn BEV layouts depicting various road configurations.
For all tested scenarios, our model generates diverse and plausible 3D semantic scenes. The red boxes highlight the layout regions, confirming that the model faithfully respects the user-specified elements.
Notably, for a given layout, each generation produces a different yet coherent completion, demonstrating the diversity of the generation process.

\paragraph{Inpainting.}
\label{subsec:inpainting}
As shown in~\cref{fig:inpainting_outpainting} (top), given a scene with a masked region, the generated content blends seamlessly with the preserved regions, producing realistic transitions without visible artifacts.

\paragraph{Outpainting.}
\label{subsec:outpainting}
\cref{fig:inpainting_outpainting} (bottom) shows outpainting results, where the model produces spatially coherent generation that naturally extends the road layout, vegetation, and surrounding structures of the input scene.

Beyond sketch-guided generation, inpainting and outpainting further demonstrate the flexibility offered by the discrete latent space of our pipeline, enabling diverse scene editing scenarios \textit{without any retraining}.
}

\subsection{Ablation Study}
\label{sec:exp_ablation}
\paragraph{VQ-VAE configuration.}

\edit{In~\cref{tab:ae_ablation}, we} ablate the VQ-VAE configuration by varying the codebook size (512, 1024, 2048) and the latent channel dimension (4, 8, 16).
We observe that larger codebooks and higher dimensions generally improve reconstruction, with the 2048 codes / dim 8 configuration achieving the best scores\raoul{; \edit{although we highlight that higher reconstruction does not always lead to higher diffusion capability, as discussed in~\cref{sec:pilot_study}.}}
Besides, we find that codebook utilization significantly drops with larger codebooks, from 100\% with 512 codes to 43.7\% with 2048 codes.
A low utilization implies that a large portion of the codebook entries are never used, resulting in wasted capacity.
\edit{Therefore, for efficiency reasons, we choose to use 512 codes with dim 8 in our final pipeline, as this configuration achieves competitive diffusion performance \textit{while ensuring full codebook utilization}}.

\begin{table}[ht]
    \centering
    \resizebox{1.0\linewidth}{!}{%
    \begin{tabular}{ccc|cc|ccc}
         \toprule
         \multicolumn{3}{c}{}& \multicolumn{2}{c}{Autoencoder}& \multicolumn{3}{c}{Diffusion}\\
         Codes & Dim. & \makecell{Codebook\\utilization $\uparrow$} & IoU$\uparrow$ & mIoU$\uparrow$ & FID$\downarrow$ & KID$\downarrow$ & CKL$\downarrow$  \\
         \midrule
         512 & 4 & & 84.00 & 71.08 & - & - & -\\
         \rowOurs{}512 & 8 & \best{100\%} & 81.90 & 72.20 & \best{84.9} & 0.0818 & 0.0362\\
         512 & 16 & & 83.90 & 72.80 & - & - & -\\
         \midrule
         1024 & 8 & 59.1\% & 84.10 & 72.50 & 85.70 & \best{0.0812} & \best{0.0326} \\
         \midrule
         2048 & 8 & 43.7\% & \best{85.09} & \best{74.04}  & 93.77 & 0.0914 & 0.0361 \\
         \bottomrule
    \end{tabular}%
    }
    \caption{\textbf{VQ-VAE ablation.} Reconstruction performance on the SemanticKITTI validation set for varying codebook sizes and latent dimensions. We select 512 codes with dimension 8, which produces reasonably good performance while ensuring full codebook utilization.}
    \label{tab:ae_ablation}
\end{table}

\paragraph{Pre vs post-quantization diffusion.}

Our training-free capability directly emerges from our choice of diffusion on discrete latents, which enables code-to-class mapping (\cf \cref{sec:meth_editing}).
We assess the adequacy of such a choice by comparing our discrete latent diffusion (\ie, after quantization) with diffusion on continuous latents (before quantization).
Results in~\cref{tab:quant_ablation} show that while the continuous variant achieves lower FID and KID, suggesting marginally better per-sample quality, discrete latent diffusion obtains better CKL and Recall, indicating a more faithful class distribution and greater diversity. Overall, we consider both variants to produce on-par performance, which validates our choice of discrete latent diffusion given its unique property.

\begin{table}[ht]
    \centering
    \resizebox{1.0\linewidth}{!}{%
    \begin{tabular}{lcccccc}
         \toprule
         Diffusion & Editable & FID$\downarrow$ & KID$\downarrow$ & CKL$\downarrow$ & Prec$\uparrow$ & Rec$\uparrow$ \\
         \midrule
         Pre-quant. latent & \ding{55} & \best{81.60} & \best{0.0777} & 0.0435 & \best{0.094} & 0.1276 \\
         \rowOurs{}Post-quant. latent & \checkmark & 84.90 & 0.0818 & \best{0.0362} & 0.086 & \best{0.1374} \\
         \bottomrule
    \end{tabular}%
    }
    \caption{\textbf{Pre- vs post-quantization diffusion.} While both variants achieve comparable overall performance, only post-quantization latent diffusion enables training-free, editable scene generation.}
    \label{tab:quant_ablation}
\end{table}

\section{Conclusion and perspective.}
\label{sec:conclusion}
In this paper, we present EditSSC, a diffusion-based pipeline for 3D semantic occupancy scene generation. By leveraging a latent Bird’s Eye View (BEV) representation, our method enables intuitive latent editing within an image-like spatial domain. While existing 3D reference methods often suffer performance degradation when adapted to a BEV setup, our approach outperforms them in unconditional generation and remains  competitive when conditioned on LiDAR frames.

The integration of a Vector Quantized (VQ) representation allows for direct manipulation of the latent space. By prescribing specific class latents within designated areas, we can generate conditioned scenes out of the box, without requiring any network retraining.

EditSSC demonstrates promising results, opening several avenues for future work: improving conditional performance while preserving the simplicity of the design; expanding training datasets to more effectively utilize the capacity of the VQ codebook or integrating pretrained diffusion models.
The BEV representation facilitates the use of standard and pretrained 2D diffusion models. This presents an appealing direction for leveraging large-scale pretraining and incorporating multi-modal conditioning, such as text to 3D scene generation.

\paragraph*{Acknowledgments}
This work was performed using HPC resources from GENCI–IDRIS (Grant AD011017284, AD011012883R4).

{
    \small
    \bibliographystyle{ieeenat_fullname}
    \bibliography{main}

@String(CVPR= {IEEE Conf. Comput. Vis. Pattern Recog.})

@String(ICCV= {Int. Conf. Comput. Vis.})

@String(ECCV= {Eur. Conf. Comput. Vis.})

@String(NIPS= {Adv. Neural Inform. Process. Syst.})

@String(TOG= {ACM Trans. Graph.})

@String(ICLR = {Int. Conf. Learn. Represent.})

@String(AAAI = {AAAI})

@String(CVPR  = {CVPR})

@String(ICCV  = {ICCV})

@String(ECCV  = {ECCV})

@String(NIPS  = {NeurIPS})

@String(ICML  = {ICML})

@String(TOG   = {ACM TOG})

@String(ICLR  = {ICLR})

@inproceedings{liao2025diffusiondrive,
  title={Diffusiondrive: Truncated diffusion model for end-to-end autonomous driving},
  author={Liao, Bencheng and Chen, Shaoyu and Yin, Haoran and Jiang, Bo and Wang, Cheng and Yan, Sixu and Zhang, Xinbang and Li, Xiangyu and Zhang, Ying and Zhang, Qian and others},
  booktitle=CVPR,
  year={2025}
}

@inproceedings{kouzelis2025eq,
  title={Eq-vae: Equivariance regularized latent space for improved generative image modeling},
  author={Kouzelis, Theodoros and Kakogeorgiou, Ioannis and Gidaris, Spyros and Komodakis, Nikos},
  booktitle=ICML,
  year={2025}
}

@inproceedings{lopes2025matswap,
  title={MatSwap: Light-aware material transfers in images},
  author={Lopes, Ivan and Deschaintre, Valentin and Hold-Geoffroy, Yannick and de Charette, Raoul},
  booktitle={Computer Graphics Forum},
  volume={44},
  number={4},
  year={2025},
  organization={Wiley Online Library}
}

@inproceedings{li2023diffusion,
  title={Diffusion-sdf: Text-to-shape via voxelized diffusion},
  author={Li, Muheng and Duan, Yueqi and Zhou, Jie and Lu, Jiwen},
  booktitle=CVPR,
  year={2023}
}

@inproceedings{maruani2025shapeshifter,
  title={ShapeShifter: 3D Variations Using Multiscale and Sparse Point-Voxel Diffusion},
  author={Maruani, Nissim and Yifan, Wang and Fisher, Matthew and Alliez, Pierre and Desbrun, Mathieu},
  booktitle=CVPR,
  year={2025}
}

@inproceedings{liu2023meshdiffusion,
  title={Meshdiffusion: Score-based generative 3d mesh modeling},
  author={Liu, Zhen and Feng, Yao and Black, Michael J and Nowrouzezahrai, Derek and Paull, Liam and Liu, Weiyang},
  booktitle=ICLR,
  year={2023}
}

@article{jun2023shap,
  title={Shap-e: Generating conditional 3d implicit functions},
  author={Jun, Heewoo and Nichol, Alex},
  journal={arXiv preprint arXiv:2305.02463},
  year={2023}
}

@inproceedings{shim2023diffusion,
  title={Diffusion-based signed distance fields for 3d shape generation},
  author={Shim, Jaehyeok and Kang, Changwoo and Joo, Kyungdon},
  booktitle=CVPR,
  year={2023}
}

@inproceedings{shue20233d,
  title={3d neural field generation using triplane diffusion},
  author={Shue, J Ryan and Chan, Eric Ryan and Po, Ryan and Ankner, Zachary and Wu, Jiajun and Wetzstein, Gordon},
  booktitle=CVPR,
  year={2023}
}

@inproceedings{luo2021diffusion,
  title={Diffusion probabilistic models for 3d point cloud generation},
  author={Luo, Shitong and Hu, Wei},
  booktitle=CVPR,
  year={2021}
}

@article{vahdat2022lion,
  title={Lion: Latent point diffusion models for 3d shape generation},
  author={Vahdat, Arash and Williams, Francis and Gojcic, Zan and Litany, Or and Fidler, Sanja and Kreis, Karsten and others},
  journal=NIPS,
  volume={35},
  year={2022}
}

@article{wu2024blockfusion,
  title={Blockfusion: Expandable 3d scene generation using latent tri-plane extrapolation},
  author={Wu, Zhennan and Li, Yang and Yan, Han and Shang, Taizhang and Sun, Weixuan and Wang, Senbo and Cui, Ruikai and Liu, Weizhe and Sato, Hiroyuki and Li, Hongdong and others},
  journal={ACM Transactions on Graphics (ToG)},
  volume={43},
  number={4},
  year={2024},
  publisher={ACM New York, NY, USA}
}

@inproceedings{meng2025lt3sd,
  title={Lt3sd: Latent trees for 3d scene diffusion},
  author={Meng, Quan and Li, Lei and Nie{\ss}ner, Matthias and Dai, Angela},
  booktitle=CVPR,
  year={2025}
}

@inproceedings{zhou20213d,
  title={3d shape generation and completion through point-voxel diffusion},
  author={Zhou, Linqi and Du, Yilun and Wu, Jiajun},
  booktitle=ICCV,
  year={2021}
}

@inproceedings{ju2024diffindscene,
  title={Diffindscene: Diffusion-based high-quality 3d indoor scene generation},
  author={Ju, Xiaoliang and Huang, Zhaoyang and Li, Yijin and Zhang, Guofeng and Qiao, Yu and Li, Hongsheng},
  booktitle=CVPR,
  year={2024}
}

@inproceedings{kim2023neuralfield,
  title={Neuralfield-ldm: Scene generation with hierarchical latent diffusion models},
  author={Kim, Seung Wook and Brown, Bradley and Yin, Kangxue and Kreis, Karsten and Schwarz, Katja and Li, Daiqing and Rombach, Robin and Torralba, Antonio and Fidler, Sanja},
  booktitle=CVPR,
  year={2023}
}

@inproceedings{liu2024pyramid,
  title={Pyramid diffusion for fine 3d large scene generation},
  author={Liu, Yuheng and Li, Xinke and Li, Xueting and Qi, Lu and Li, Chongshou and Yang, Ming-Hsuan},
  booktitle=ECCV,
  year={2024},
  organization={Springer}
}

@inproceedings{ren2024xcube,
  title={Xcube: Large-scale 3d generative modeling using sparse voxel hierarchies},
  author={Ren, Xuanchi and Huang, Jiahui and Zeng, Xiaohui and Museth, Ken and Fidler, Sanja and Williams, Francis},
  booktitle=CVPR,
  year={2024}
}

@inproceedings{bahmani2023cc3d,
  title={Cc3d: Layout-conditioned generation of compositional 3d scenes},
  author={Bahmani, Sherwin and Park, Jeong Joon and Paschalidou, Despoina and Yan, Xingguang and Wetzstein, Gordon and Guibas, Leonidas and Tagliasacchi, Andrea},
  booktitle=ICCV,
  year={2023}
}

@inproceedings{semcity,
  title={{Semcity: Semantic scene generation with triplane diffusion}},
  author={Lee, Jumin and Lee, Sebin and Jo, Changho and Im, Woobin and Seon, Juhyeong and Yoon, Sung-Eui},
  booktitle=CVPR,
  year={2024}
}

@article{ho2022classifier,
  title={Classifier-free diffusion guidance},
  author={Ho, Jonathan and Salimans, Tim},
  journal={arXiv preprint arXiv:2207.12598},
  year={2022}
}

@article{scenescalediff,
  title={Diffusion Probabilistic Models for Scene-Scale 3D Categorical Data},
  author={Lee, Jumin and Im, Woobin and Lee, Sebin and Yoon, Sung-Eui},
  journal={arXiv preprint arXiv:2301.00527},
  year={2023}
}

@inproceedings{semantickitti,
  title={Semantickitti: A dataset for semantic scene understanding of lidar sequences},
  author={Behley, Jens and Garbade, Martin and Milioto, Andres and Quenzel, Jan and Behnke, Sven and Stachniss, Cyrill and Gall, Jurgen},
  booktitle=ICCV,
  year={2019}
}

@inproceedings{rombach2022high,
  title={High-resolution image synthesis with latent diffusion models},
  author={Rombach, Robin and Blattmann, Andreas and Lorenz, Dominik and Esser, Patrick and Ommer, Bj{\"o}rn},
  booktitle=CVPR,
  year={2022}
}

@misc{oquab2023dinov2,
  title={DINOv2: Learning Robust Visual Features without Supervision},
  author={Oquab, Maxime and Darcet, Timothée and Moutakanni, Theo and Vo, Huy V. and Szafraniec, Marc and Khalidov, Vasil and Fernandez, Pierre and Haziza, Daniel and Massa, Francisco and El-Nouby, Alaaeldin and Howes, Russell and Huang, Po-Yao and Xu, Hu and Sharma, Vasu and Li, Shang-Wen and Galuba, Wojciech and Rabbat, Mike and Assran, Mido and Ballas, Nicolas and Synnaeve, Gabriel and Misra, Ishan and Jegou, Herve and Mairal, Julien and Labatut, Patrick and Joulin, Armand and Bojanowski, Piotr},
  journal={arXiv:2304.07193},
  year={2023}
}

@inproceedings{yu2024repa,
  title={Representation alignment for generation: Training diffusion transformers is easier than you think},
  author={Yu, Sihyun and Kwak, Sangkyung and Jang, Huiwon and Jeong, Jongheon and Huang, Jonathan and Shin, Jinwoo and Xie, Saining},
  booktitle=ICLR ,
  year={2024}
}

@inproceedings{berman2018lovasz,
  title={The lov{\'a}sz-softmax loss: A tractable surrogate for the optimization of the intersection-over-union measure in neural networks},
  author={Berman, Maxim and Triki, Amal Rannen and Blaschko, Matthew B},
  booktitle=CVPR,
  year={2018}
}

@article{ho2020denoising,
  title={Denoising diffusion probabilistic models},
  author={Ho, Jonathan and Jain, Ajay and Abbeel, Pieter},
  journal=NIPS,
  volume={33},
  year={2020}
}

@inproceedings{lugmayr2022repaint,
  title={Repaint: Inpainting using denoising diffusion probabilistic models},
  author={Lugmayr, Andreas and Danelljan, Martin and Romero, Andres and Yu, Fisher and Timofte, Radu and Van Gool, Luc},
  booktitle=CVPR,
  year={2022}
}

@inproceedings{saharia2022palette,
  title={Palette: Image-to-image diffusion models},
  author={Saharia, Chitwan and Chan, William and Chang, Huiwen and Lee, Chris and Ho, Jonathan and Salimans, Tim and Fleet, David and Norouzi, Mohammad},
  booktitle={ACM SIGGRAPH 2022 conference proceedings},
  year={2022}
}

@article{saharia2022photorealistic,
  title={Photorealistic text-to-image diffusion models with deep language understanding},
  author={Saharia, Chitwan and Chan, William and Saxena, Saurabh and Li, Lala and Whang, Jay and Denton, Emily L and Ghasemipour, Kamyar and Gontijo Lopes, Raphael and Karagol Ayan, Burcu and Salimans, Tim and others},
  journal=NIPS,
  volume={35},
  year={2022}
}

@inproceedings{tang2024diffuscene,
  title={Diffuscene: Denoising diffusion models for generative indoor scene synthesis},
  author={Tang, Jiapeng and Nie, Yinyu and Markhasin, Lev and Dai, Angela and Thies, Justus and Nie{\ss}ner, Matthias},
  booktitle=CVPR,
  year={2024}
}

@article{heusel2017gans,
  title={Gans trained by a two time-scale update rule converge to a local nash equilibrium},
  author={Heusel, Martin and Ramsauer, Hubert and Unterthiner, Thomas and Nessler, Bernhard and Hochreiter, Sepp},
  journal=NIPS,
  volume={30},
  year={2017}
}

@inproceedings{binkowski2018demystifying,
  title={Demystifying mmd gans},
  author={Bi{\'n}kowski, Miko{\l}aj and Sutherland, Danica J and Arbel, Michael and Gretton, Arthur},
  booktitle=ICLR,
  year={2018}
}

@article{vqvae,
  title={Neural discrete representation learning},
  author={Van Den Oord, Aaron and Vinyals, Oriol and others},
  journal=NIPS,
  volume={30},
  year={2017}
}

@inproceedings{ts3d,
  title={Two stream 3d semantic scene completion},
  author={Garbade, Martin and Chen, Yueh-Tung and Sawatzky, Johann and Gall, Juergen},
  booktitle={CVPR-W},
  year={2019}
}

@inproceedings{lmscnet,
  title={{LMSCNet}: Lightweight multiscale {3D} semantic completion},
  author={Roldao, Luis and De Charette, Raoul and Verroust-Blondet, Anne},
  booktitle={3DV},
  year={2020},
  xxxorganization={IEEE}
}

@inproceedings{js3cnet,
  title={Sparse Single Sweep LiDAR Point Cloud Segmentation via Learning Contextual Shape Priors from Scene Completion},
  author={Yan, Xu and Gao, Jiantao and Li, Jie and Zhang, Ruimao and Li, Zhen and Huang, Rui and Cui, Shuguang},
  booktitle={AAAI},
  year={2021}
}

@inproceedings{diffssc,
  title={DiffSSC: Semantic LiDAR scan completion using denoising diffusion probabilistic models},
  author={Cao, Helin and Behnke, Sven},
  booktitle={IROS},
  year={2025},
}

@article{liu5333789dual,
  title={A Dual-Path Network for Semantic Scene Completion of Single-Frame Lidar Point Clouds},
  author={Liu, Wei and Kang, Ziwen and Yu, Yongtao and Gong, Zheng and Zheng, Yuchao and Huang, Xiaohui and Guan, Haiyan and Ma, Lingfei and Zhang, Dedong},
  journal={Int. J. Appl. Earth Obs.},
  year={2026}
}

@article{xi2026flowssc,
  title={FlowSSC: Universal Generative Monocular Semantic Scene Completion via One-Step Latent Diffusion},
  author={Xi, Zichen and Chen, Hao-Xiang and Xue, Nan and Yan, Hongyu and Feng, Qi-Yuan and Kara, Levent Burak and Jorge, Joaquim and Xu, Qun-Ce},
  journal={Robotics and Automation Letters},
  year={2026}
}

@article{zheng2026sseditor,
  title={Sseditor: Controllable mask-to-scene generation with diffusion model},
  author={Zheng, Haowen and Pang, Jiahao and Pu, Zhiqiang and Liang, Yanyan},
  journal={Knowledge-Based Systems},
  year={2026},
  publisher={Elsevier}
}

@inproceedings{esser2024scaling,
  title={Scaling rectified flow transformers for high-resolution image synthesis},
  author={Esser, Patrick and Kulal, Sumith and Blattmann, Andreas and Entezari, Rahim and M{\"u}ller, Jonas and Saini, Harry and Levi, Yam and Lorenz, Dominik and Sauer, Axel and Boesel, Frederic and others},
  booktitle=ICML,
  year={2024}
}

@inproceedings{lee2025latent,
  title={Latent diffusion models with masked autoencoders},
  author={Lee, Junho and Shin, Jeongwoo and Choi, Hyungwook and Lee, Joonseok},
  booktitle=ICCV,
  year={2025}
}
}


\end{document}